\documentclass[11pt]{article}
\usepackage{coling2020}
\usepackage{times}
\usepackage{url}
\usepackage{latexsym}

\usepackage{graphicx}
\usepackage{array}

\usepackage{caption}
\usepackage{subcaption}

\colingfinalcopy 

\title{``What is on your mind?'' \\ Automated Scoring of Mindreading in Childhood and Early Adolescence}

\author{Venelin Kovatchev\textsuperscript{1}, Phillip Smith\textsuperscript{2}, Mark Lee\textsuperscript{2}, Imogen Grumley Traynor\textsuperscript{1}, \\ \textbf{Irene Luque Aguilera\textsuperscript{1} and Rory T. Devine\textsuperscript{1}} \\ 
\textsuperscript{1} School of Psychology, University of Birmingham \\
\textsuperscript{2} School of Computer Science, University of Birmingham \\
52 Prichatts Road, Edgbaston, B15 2SB, United Kingdom \\
\{v.o.kovatchev , P.Smith.7 , m.g.lee , I.GrumleyTraynor , \\ I.LuqueAguilera , R.T.Devine\} @bham.ac.uk
}

\date{}

\begin{document}
\maketitle
\begin{abstract}
In this paper we present the first work on the automated scoring of mindreading ability in middle childhood and early adolescence. We create MIND-CA, a new corpus of 11,311 question-answer pairs in English from 1,066 children aged 7 to 14. We perform machine learning experiments and carry out extensive quantitative and qualitative evaluation. We obtain promising results, demonstrating the applicability of state-of-the-art NLP solutions to a new domain and task.

\end{abstract}

\section{Introduction}
\label{intro}

\blfootnote{    
\hspace{-0.65cm}  
This work is licensed under a Creative Commons 
Attribution 4.0 International Licence.
\\ Licence details:
\url{http://creativecommons.org/licenses/by/4.0/}.
}
Successfully navigating day-to-day social interactions often hinges on being able to recognize that others can hold different beliefs, thoughts, ideas, intentions and desires. Popular culture and literature are replete with instances that require us to enter into the mind of someone else in order to explain their actions. For example, the final scene of Shakespeare's Romeo and Juliet is made all the more poignant because the audience know that Juliet is still alive and that Romeo has killed himself because he had a mistaken belief that she was dead. The audience is able to understand both the state of the world (i.e.: Juliet is still alive) and the mistaken belief of Romeo (i.e.: Juliet is dead).

The ability to attribute mental states to others in order to explain or predict behavior, called ``theory of mind'' or ``mindreading'' has intrigued researchers in disciplines as diverse as psychology, philosophy, neuroscience,
economics, and literary theory for four decades \cite{HughesDevine2015}. The long-standing interest in mindreading is not surprising as children's performance on standard tests of mindreading has real-world significance. Longitudinal studies demonstrate that children who excel at tests of mindreading are more likely than their peers to be identified as popular by classmates \cite{doi:10.1111/j.1467-8624.2011.01669.x}, have reciprocated friendships \cite{Fink}, and be viewed by their teachers as socially competent \cite{PMID:26914214}. Furthermore, case-control studies show that deficits in mindreading are more common among individuals with a range of mental health problems and neuro-developmental conditions associated with poorer social functioning \cite{COTTER201892}. Given the prognostic importance of mindreading in children, in this paper we present the joint work of an inter-disciplinary team of psychologists, computer scientists, and linguists on the automatic scoring of mindreading ability in middle childhood and early adolescence.

The great majority of studies on mindreading have focused on identifying the age at which preschool children recognize that others have desires, beliefs and knowledge that differ from their own \cite{doi:10.1111/j.1467-8624.2004.00691.x}. However, the past decade has witnessed increased interest in the continued growth of children’s understanding of mind in middle childhood and early adolescence \cite{sfilm,doi:10.1111/j.1467-7687.2009.00888.x,lagattuta}. There is currently no consensus on how best to measure mindreading in middle childhood and adolescence but standardized open-ended tests in which children are asked to explain the behavior of a character depicted in short vignettes \cite{happe,doi:10.1111/j.1467-8624.2011.01669.x} or in animations or film clips \cite{castelli,sfilm} are increasingly used. While promising, open-ended responses to  test questions are currently manually rated by trained experts limiting the potential utility of these measures in large-scale research. This challenge is not unique to research on mindreading as open-ended text response assessments are widely-used in clinical and educational research \cite{heetal,lee2019,zehner}. Automated scoring systems have the potential to enhance reliability and minimize the amount of labor required to administer tests at scale. 

To address this problem, we gathered, digitalized, and manually rated responses to two tests of mindreading: the Strange Stories task \cite{happe} and the Silent Film task \cite{sfilm}. The new corpus, that we call MIND-CA, contains  11,311 open ended responses in English from children aged 7 to 14. We used MIND-CA to train and evaluate several different Machine Learning systems. We performed an extensive quantitative and qualitative evaluation and error analysis to measure the applicability of the automated systems. 

Our findings demonstrate that NLP can be used to aid the work of psychologists, educational researchers, clinicians and psycho-linguists. State-of-the-art text classification systems report very good quantitative results when scoring children's mindreading capabilities. At the same time, the in-depth analysis of the performance indicates that there is still room for improvement and future work.


The rest of this article is organized as follows. Section \ref{wmr:related} presents related work. Section \ref{wmr:data} provides details about the creation of the dataset. Section \ref{wmr:setup} describes the experimental setup and the evaluated systems. Section \ref{wmr:results} presents the results of the automatic scoring and the qualitative error analysis. Section \ref{wmr:discussion} discusses the implications of the findings of this work. Finally, Section \ref{wmr:conclusions} presents the conclusions and outlines directions for future research.

\section{Related Work}\label{wmr:related}

The extension of research on mindreading into middle childhood and adolescence has stimulated the creation of new tests because the standard tasks used to measure mindreading in early childhood (e.g., the false belief task) exhibit marked ceiling effects by age 6 \cite{wellman2001}. Advanced tests of mindreading used in middle childhood and adolescence employ a range of stimuli (e.g., vignettes, animations, film clips) and appear to target distinct skills (e.g., perspective taking, intention attribution, explaining behavior) \cite{doi:10.1111/j.1467-8624.2011.01669.x,doi:10.1111/j.1467-7687.2009.00888.x,castelli}. Despite these in-roads, research with school-aged children has revealed limited convergence between tasks (i.e., suggested tasks may not capture the same underlying ability) \cite{WARNELL2019103997} and limited discriminating power in some tasks (i.e., suggested tasks may not capture sufficient variance in middle childhood) \cite{COOPER201929,doi:10.1111/cdev.12566}. 

	In contrast with these results, studies of children aged between 7 and 13 years have shown consistent associations between children’s performance on two tests: the Strange Stories \cite{happe} and the Silent Film task \cite{sfilm}. The Strange Stories task was developed for studying mindreading in adolescents and adults with autism. In this task participants answer questions about characters in short vignettes depicting instances of mistaken beliefs, misunderstanding, double-bluff or deception. Supporting its validity, adolescents and adults with autism (a condition associated with impaired social and communicative ability) were less likely than 'neurotypical' controls to explain characters' behaviours with reference to their thoughts, feelings and desires \cite{white2009}. Building on this work, \newcite{sfilm} devised the Silent Film task in an effort to create a measure of individual differences in mindreading suitable for use in middle childhood and adolescence. Like the Strange Stories task, the Silent Film task requires participants to answer open-ended questions about a character's behavior with reference to the character's mental states but uses brief clips taken from a classic silent comedy. The two tests show convergent validity in middle childhood and early adolescence and marked individual differences in performance in each age group studied \cite{sfilm,DevineHughes2016}. The two tasks exhibit strong one-month test-retest reliability \cite{DevineHughes2016} and the validity of these tasks is bolstered by evidence of correlations between test performance and measures of children's social competence in middle childhood and early adolescence \cite{PMID:26914214}. 

The Strange Stories and Silent Film tasks are promising measures with robust psychometric properties. However, like other open-ended measures in psychology, children's responses to the test questions on the Strange Stories and Silent Film tasks are manually rated by trained researchers limiting the potential utility of these measures in large-scale research for two reasons. First, manual coding of open-ended text responses (even short responses) is labor-intensive and time-consuming. Second, even trained researchers may change the way they apply scoring rubrics over time or make errors in applying a scoring rubric. Such ``drift'' is not detected in standard inter-rater agreement assessments and potentially undermines the reliability of manual ratings \cite{kazdin}. The creation of an automated scoring platform for the Strange Stories and Silent Film tasks raises the possibility of rapid, reliable assessment of individual differences in mindreading ability across middle childhood and adolescence.

Automated scoring is not a novel application within NLP. It is usually deployed in the educational domain. \newcite{burrows}, \newcite{zesch-etal-2015-task}, and \newcite{sultan-etal-2016-fast} present different work on automated scoring. The ASAP\footnote{\url{https://www.kaggle.com/c/asap-aes}} and ASAP2\footnote{\url{https://www.kaggle.com/c/asap-sas}} as well as the SemEval 2013 shared task 7 \cite{dzikovska-etal-2013-semeval} helped popularize the topic within NLP researchers. While the majority of the work is focused on essay grading, \newcite{zehner} implement and integrate baselines of NLP for the automated coding of short text responses. \newcite{lee2019} describe an automated scoring and feedback system that can help students revise and improve their scientific argumentation. 

\newcite{madnani-cahill-2018-automated} present a position paper about using NLP for automated scoring. They argue that automatic scoring is a complex problem with multiple perspectives, and “stakeholder” implications. As such, they suggest that designing a successful automated scoring solution requires collaboration between NLP researchers, subject matter experts, and score users. They also point out that the nature and importance of automated scoring requires a much more thorough evaluation and analysis. 

In this paper we present the first work on the automated scoring of children's open-ended responses to tests of mindreading ability. We follow the best practices in automated scoring in order to ensure the applicability and quality of the proposed solution. 

\section{Dataset Creation}\label{wmr:data}

In this section, we present the process of creating a novel NLP resource - the MIND-CA corpus. Section \ref{wmr:tasks} describes in more details the two mindreading tests that we use: the \textit{Strange Stories Task} and the \textit{Silent Film Task}. Section \ref{wmr:dataset} details the corpus creation process, the format of the final corpus, as well as some statistical information. 

\subsection{Strange Story Task and Silent Film Task}\label{wmr:tasks}

In the \textit{Strange Stories Task} \cite{happe} children listen to the researcher read five vignettes depicting different social situations (e.g., lying, misunderstanding, double bluffing) while the story text remains visible on a large screen. Each story is followed by a single open-response question in which the children are required to explain a character’s behavior with reference to the character’s mental states. 

In the \textit{Silent Film Task} \cite{sfilm} children watch five short film clips on a large screen. The clips were taken from a classic silent comedy and depicted instances of deception, misunderstanding, and false belief. Children watch each clip once and respond to a question read aloud by the researcher after each clip (the first clip is followed by two questions). Children record their responses in writing. 

Children's performance on the two tasks is scored manually by trained experts\footnote{The full guidelines for scoring are available at \url{https://github.com/venelink/mindreading-coling}}. The scoring scheme for each item on the two tasks has three possible grades. \textbf{Score 0} - \textit{failure to mindread or inappropriate mindreading given the context}; \textbf{Score 1} - \textit{reference to mind or mental state but partially inappropriate explanation/interpretation}; \textbf{Score 2} - \textit{reference to mind or mental state with appropriate explanation/interpretation}. Scores across all items are then used to create total scores to capture individual differences in mindreading \cite{sfilm}. 

\subsection{The MIND-CA Corpus}\label{wmr:dataset}

The dataset creation process has three steps: 
1) gathering children's responses; 
2) digitalization and scoring of responses; 
3) anonymization, reformatting, and pre-processing the data for NLP purposes.

\textbf{Gathering children responses} We recruited 1,066 English-speaking children aged between 7.25 and 13.53 years, M\textsubscript{age} = 10.32 years, SD = 1.43, from 46 classrooms in 13 primary and 4 secondary schools in England between 2014 and 2019. The children took part in a whole-class testing session lasting approximately 1 hour led by a trained research assistant using a scripted protocol \cite{DevineHughes2016}. Each child answered 11 questions - five in the \textit{Strange Story Task} and six in the \textit{Silent Film Task}. We obtained a total of 11,726 question-answer pairs.

\textbf{Digitalization and scoring} The paper test booklets were digitalized and manually scored by two post-graduate research assistants and the test developer. Each booklet was scored by one research assistant and difficult cases were discussed during weekly meetings. To ensure the quality and consistency of the scoring, the annotators were trained for the task on a held out set of 30 cases for each task and reported an inter-annotator agreement with the test developer of 94.4\% and a Fleiss' Kappa of .91. The inter-annotator agreement was measured again on a second held-out set at the end of the annotation, and the results were consistent with the training. During the digitalization of children's written responses, the annotators were instructed to preserve the original spelling and grammar where possible. We also created a dictionary of the most common spelling mistakes.

\textbf{Corpus preprocessing} After all responses were digitalized and scored, we anonymized them by removing any personal information that can be used to identify the participants. We kept only the information considered to be relevant for the rating and the analysis of the results. We used the spaCy Python library \cite{spacy2} to run basic linguistic pre-processing of the corpus (tokenization, part of speech tagging) and the NLTK Python library \cite{BirdKleinLoper09} to obtain basic corpus statistics. 

\textbf{Final corpus format} Table \ref{wmr:corp-samp-r} shows a sample of the final corpus. It contains the following columns: 1) the question; 2) the open-text response; 3) the gold-standard score; 4) the age of the participant; 5) the gender of the participant. We created a second version of the corpus where we explicitly incorporated the question in the ``Response'' column. That is, in the ``full'' version of the corpus, the ``Response'' column of the first entry in Table \ref{wmr:corp-samp-r} is \textit{``Why did the men hide ? Because they did n't want the woman to find them''}. All other columns remain the same. In our experiments we use both versions of the corpus\footnote{The corpora is publicly available at \url{https://github.com/venelink/mindreading-coling}} and compare the performance of the automated systems. 

\begin{table}[h!]
\begin{center}
\begin{tabular}{|l|p{7.5cm}|r|r|r|}
\hline
Question & Response & Score & Age & Gender \\ \hline
Why did the men hide? &  Because they did n't want the woman to find them & 2.0 & 9 & 0 \\ \hline
Why did the men hide? & Because they Were n't supose to be there & 0.0 & 9 & 1 \\ \hline  
Why did the men hide? & they have a seccret & 0.0 & 9 & 0 \\ \hline 
Why did the men hide? & because they didnt whant to get cort & 1.0 & 9 & 0 \\ \hline 
Why did the men hide? & Because they did n't want the woman to see them & 1.0 & 9 & 1 \\ \hline   

\end{tabular}
\end{center}
\caption{Examples from the Corpus MIND-CA}
\label{wmr:corp-samp-r}
\end{table}

\textbf{Corpus statistics} After filtering unreadable and missing data, the final corpus contains 11,311 question-answer pairs from 1,048 unique participants. In terms of gender 5,682 of the question-answer pairs are from girls, 5,596 of the answers are from boys. The age of the participants ranges from 7 to 14, and the distribution is as follows: 102 answers from participant at the age of 7; 668 at the age of 8; 2,809 at the age of 9; 3,061 at the age of 10; 2,163 at the age of 11; 1,333 at the age of 12; 1,142 at the age of 13; and 33 at the age of 14.  The distribution of labels in the full corpus is well balanced: 3,517 pairs (31 \%) with a score of 0; 3,822 pairs (33.8 \%) with a score of 1; and 3,972 pairs (35.1 \%) with a score of 2. The average length of responses (in tokens) is 10.1, and it varies per question from 8.25 to 11.64.

\section{Experimental Setup}\label{wmr:setup}

The main objective of this work was to create and evaluate an automated solution for scoring children's mindreading ability from open-ended responses to two standardized tests. A successful solution will facilitate large scale psychological research on mindreading. We formally defined the task of automatic scoring as a classification problem. We chose classification over regression taking in account the specifics of the original tasks and the data, and the need to have item-level scores for future research. 
For our experimental setup, we selected four different machine learning architectures. The systems were chosen based on two main criteria: 1) they represent qualitatively different approaches to the formalization and processing of the data; 2) they are sufficiently lightweight in terms of computational requirements to be implemented in a web-based end-to-end application for researchers in psychology. The systems are:

\begin{itemize}
    \item An SVM classifier, trained with basic linguistic features (bag of words, bag of word ngrams, bag of character ngrams, part of speech frequencies). We tested both raw frequencies and tf-idf weighting.
    \item A Bi-LSTM neural network.
    \item A self-attention based neural network. 
    \item A pre-trained DistilBERT transformer, finetuned on the dataset.
\end{itemize}

The systems were implemented using Scikit-Learn \cite{scikit-learn}, TensorFlow 2.0 \cite{tensorflow2015-whitepaper}, HuggingFace \cite{Wolf2019HuggingFacesTS}, and KTrain \cite{maiya2020ktrain} libraries for Python\footnote{The code for all experiments is available at \url{https://github.com/venelink/mindreading-coling}}. All models were trained on a single GPU.
For each of the systems we tested both formats of the input described in Section \ref{wmr:dataset}. In the \textit{response-only} setup we only provided the systems with the open-ended response and the corresponding score. In the \textit{full} setup the ``Response'' column also included the question.
For the Bi-LSTM and the self-attention based models, we tried two different ways of initialization: random initialization and initialization with pre-trained FastText embeddings \cite{bojanowski2016enriching}. FastText performed much better in the first few training epochs, however after the 15th epoch the performance of the two initializations is almost identical. We report the results obtained from the random initialization.

\section{Experimental Results}\label{wmr:results}

We performed a 10-fold cross validation on the data. We further split each training set to 75\% training and 25\% validation. We report the average performance across the 10 runs for each system.

We evaluated the performance of the automated systems in four different ways. In Section \ref{wmr:acc} we report the overall performance of the systems, measured in Accuracy and Macro-F1. In Section \ref{wmr:acc-by-q} we discuss the performance of the systems on scoring each of the 11 different questions. In Section \ref{wmr:acc-by-a} we analyze the variance of the performance of the systems in terms of the age of the participants. Finally, in Section \ref{wmr:error} we present a human analysis of the incorrect predictions made by two of the systems. 

\subsection{Overall Performance}\label{wmr:acc}

\begin{table}[h!]
\begin{center}
\begin{tabular}{|l||r|r||r|r|}
\hline
System Name     & \multicolumn{2}{|c||}{Response-Only} & \multicolumn{2}{|c|}{Full} \\ \hline
            &  Acc  & Macro-F1    & Acc   & Macro-F1 \\ \hline
Majority Baseline    & .35   & .33   & .35   & .33 \\ \hline
SVM         & .69   & .69   & .71   & .71 \\ \hline
BiLSTM      & .81   & .81   & .83   & .83 \\ \hline
Attention   & .81   & .81   & .83   & .83 \\ \hline
DistilBERT  & .89   & .89   & .91   & .91 \\ \hline
\end{tabular}
\caption{Overall System Performance on the Full Test Set}
\label{wmr:tab:acc}
\end{center}

\end{table}

Table \ref{wmr:tab:acc} shows the performance of each system on the full test dataset. For each system we report the Accuracy and the Macro-F1 score. We also include a majority baseline, that is, predicting the most frequent class.
All the models obtained very good results and improved substantially over the majority baseline. Furthermore, the three deep learning models outperformed the SVM by over .10. The best performing system was DistilBERT (.89 accuracy in ``response-only'' and .91 accuracy in ``full''). Adding the questions to the input increased the accuracy of all systems by .2 - .3. When evaluating the systems on the full corpus, we found no difference between the Accuracy and the Macro-F1 scores. These quantitative results demonstrate that the problem of automatically evaluating children's mindreading can successfully be framed and processed as an NLP task. 


\subsection{Performance by Question}\label{wmr:acc-by-q}

After obtaining high overall performance, we were interested in evaluating the consistency of the systems across the 11 questions. Table \ref{wmr:tab:questions} shows a list of the question, their code (F1--F6 for Silent Film and S1--S5 for Strange Story) as well as some context for each question.

\begin{table}[h!]
\begin{center}
    
\begin{tabular}{|l|l|p{9.5cm}|}
\hline
Question & Code & Context \\ \hline
Silent film 1 & F1 & Recognize that character intends to deceive another character \\ \hline
Silent film 2 & F2 & Attribute mistaken belief to character \\ \hline
Silent film 3 & F3 & Attribute ignorance/lack of knowledge to a character \\ \hline
Silent film 4 & F4 & Attribute emotion to character based on character’s previous mistaken belief \\ \hline
Silent film 5 & F5 & Recognize character’s mistaken belief \\ \hline
Silent film 6 & F6 & Recognize that one character intends to deceive another character \\ \hline \hline
Strange story ``Brian'' & S1 & Recognize character is attempting to deceive another character \\ \hline
Strange story ``Peabody'' & S2 & Recognize one character’s mistaken beliefs about another character’s intentions \\ \hline
Strange story ``Prisoner'' & S3 & Recognize double bluff \\ \hline
Strange story ``Simon''  & S4 & Recognize awareness of deception \\ \hline
Strange story ``Burglar'' & S5 & Recognize character holds mistaken belief about another character’s beliefs \\ \hline

\end{tabular}
\caption{Questions from the Strange Story Task and the Silent Film Task}
\label{wmr:tab:questions}

\end{center}

\end{table}

Table \ref{wmr:tab:res-q} shows the performance of each system on the portion of the test set that corresponds to each of the 11 questions. Table \ref{wmr:tab:acc-q} shows the performance in terms of Accuracy, and Table \ref{wmr:tab:f1-q} shows the performance in terms of Macro-F1. To obtain the performance per question, we trained the system on the full train set and we only evaluated on the portion corresponding to the particular question. We report the average for the 10 runs in the 10-fold cross validation. For reference we also include the performance on the full test set and the simple average of all 11 questions. The success of this methodology for in-depth evaluation was demonstrated by \newcite{kovatchev-etal-2019-qualitative}.

Performance of the systems on the different questions varies both in terms of Accuracy and Macro-F1. Furthermore, while on the full test set there is no difference between Accuracy and Macro-F1, the Macro-F1 score is much lower than Accuracy when evaluating per question. Further analysis of the data revealed that while the three classes have roughly the same frequency in the full corpus, their distribution is not uniform across the different questions.
In order to determine whether the variance of the system performance was statistically significant, we carried out a Friedman-Nemenyi test \cite{Friedman:1940,Nemenyi:1963,Demsar} as implemented by the Autorank package for Python \cite{Herbold2020}. We used the Accuracy scores to perform a statistical analysis of k = 11 different questions and N = 80 paired samples (4 systems, 2 input formats, 10 runs per system). The Friedman test showed a statistically significant difference between the median values of the questions at \textit{p} \textless 0.001.

\begin{table}[h!]
\begin{center}
    \begin{subtable}{1\textwidth}
\caption{System Performance by Question (Accuracy)}
\begin{tabular}{|l||l||l||l|l|l|l|l|l||l|l|l|l|l|}
\hline
System Name             & All & Avg & \multicolumn{6}{|c||}{Silent Film Task} & \multicolumn{5}{|c|}{Strange Story Task} \\ \hline
& & & F1 & F2 & F3 & F4 & F5 & F6 & S1 & S2 & S3 & S4 & S5   \\ \hline \hline
SVM (r-only)            & \textbf{.69} & .69 & .65 & .67 & .73 & .69 & .78 & .56 & .78 & .80 & .61 & .78 & .54  \\ \hline 
BiLSTM (r-only)         & \textbf{.81} & .81 & .81 & .77 & .86 & .78 & .84 & .75 & .84 & .84 & .80 & .85 & .79 \\ \hline
Attention (r-only)      & \textbf{.81} & .81 & .80 & .78 & .84 & .80 & .83 & .76 & .85 & .85 & .79 & .85 & .75 \\ \hline
DistilBERT (r-only)     & \textbf{.89} & .89 & .88 & .86 & .91 & .86 & .88 & .84 & .92 & .92 & .85 & .92 & .89 \\ \hline \hline
SVM (full)              & \textbf{.71} & .71 & .71 & .65 & .70 & .68 & .72 & .61 & .81 & .81 & .63 & .73 & .74  \\ \hline
BiLSTM (full)           & \textbf{.83} & .83 & .80 & .76 & .87 & .80 & .87 & .77 & .85 & .87 & .82 & .85 & .83 \\ \hline
Attention (full)        & \textbf{.83} & .83 & .81 & .76 & .86 & .81 & .85 & .77 & .86 & .89 & .80 & .86 & .86  \\ \hline
DistilBERT (full)       & \textbf{.91} & .91 & .93 & .88 & .94 & .89 & .91 & .89 & .94 & .94 & .90 & .93 & .93  \\ \hline
\end{tabular}
\label{wmr:tab:acc-q}
    \end{subtable}
    
    \vspace{5mm}
    
    \begin{subtable}{1\textwidth}
\caption{System Performance by Question (F1)}
\begin{tabular}{|l||l||l||l|l|l|l|l|l||l|l|l|l|l|}
\hline
System Name             & All & Avg & \multicolumn{6}{|c||}{Silent Film Task} & \multicolumn{5}{|c|}{Strange Story Task} \\ \hline
& & & F1 & F2 & F3 & F4 & F5 & F6 & S1 & S2 & S3 & S4 & S5   \\ \hline \hline
SVM (r-only)            & \textbf{.69} & .57 & .40 & .49 & .57 & .63 & .71 & .50 & .51 & .65 & .54 & .69 & .51 \\ \hline 
BiLSTM (r-only)         & \textbf{.81} & .73 & .66 & .69 & .78 & .75 & .79 & .71 & .65 & .69 & .77 & .78 & .75 \\ \hline
Attention (r-only)      & \textbf{.81} & .72 & .65 & .69 & .77 & .77 & .79 & .72 & .67 & .68 & .75 & .77 & .71 \\ \hline
DistilBERT (r-only)     & \textbf{.89} & .83 & .83 & .80 & .86 & .84 & .85 & .80 & .84 & .75 & .82 & .87 & .87  \\ \hline \hline
SVM (full)              & \textbf{.71} & .49 & .34 & .38 & .47 & .54 & .51 & .43 & .37 & .62 & .55 & .53 & .68 \\ \hline
BiLSTM (full)           & \textbf{.83} & .73 & .59 & .67 & .81 & .73 & .83 & .73 & .62 & .70 & .78 & .70 & .80 \\ \hline
Attention (full)        & \textbf{.83} & .73 & .62 & .66 & .80 & .74 & .81 & .73 & .63 & .73 & .76 & .74 & .82 \\ \hline
DistilBERT (full)       & \textbf{.91} & .87 & .90 & .82 & .90 & .86 & .89 & .87 & .84 & .80 & .88 & .88 & .91 \\ \hline
\end{tabular}
\label{wmr:tab:f1-q}
    \end{subtable}
\end{center}
\caption{Comparing the System Performance by Question}
\label{wmr:tab:res-q}
\end{table}

\begin{figure}[h!]
\begin{center}
    
\includegraphics[width=14cm]{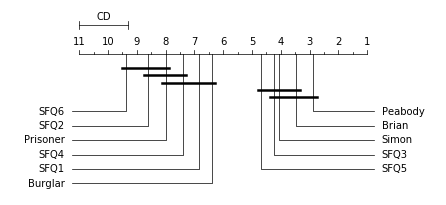}
\end{center}

\caption{Critical Difference Diagram of the Average Rank of Each Question}
\label{wmr:fn-questions}
\end{figure}

Figure \ref{wmr:fn-questions} shows the Critical Difference (CD) diagram of the post-hoc Nemenyi test. Each question is plotted with its average rank. Questions that are not connected with a horizontal line have a statistically significant difference between their median rank. From the figure it is evident that the there was a significant difference between two groups of questions.
\textbf{Easy-to-score} Questions: Silent Film questions \#3 and \#5 and Strange Story questions "Brian", "Peabody", and "Simon" were consistently easier for the automated systems to score correctly.
\textbf{Hard-to-score} Questions: Silent Film Questions \#1, \#2, \#4, and \#6 and Strange Story questions "Prisoner" and "Burglar" were consistently more challenging for the automated systems to score correctly.

The difference between \textbf{easy-to-score} and \textbf{hard-to-score} questions can also be seen in Table \ref{wmr:tab:res-q}. The three hardest questions (SFQuestion\_2 -- F2, SFQuestion\_6 -- F6, SS\_Prisoner -- S3) obtained consistently lower than average accuracy for all systems, ranging from .3 to .8 lower than \textbf{All}. The three easiest questions ("Brian" (S1), "Peabody" (S2), and "Simon" (S4)) obtained consistently higher accuracy than \textbf{All}, ranging from .2 to .4. Our annotators confirmed that the same questions were easier or more difficult to score for human experts as well.

\subsection{Performance by Age}\label{wmr:acc-by-a}

The Standards for 2014 Educational and Psychological Testing require that we ``report evidence of the variety of score interpretations for relevant subgroups in the intended population''. In the current study, the participants were grouped in terms of their age.
Table \ref{wmr:tab:acc-a} shows the performance of each system on the portion of the test set that corresponds to participants of each age group. Unlike the performance per question, in the performance per age we did not find significant differences between Accuracy and F1, so we only report the performance measured in Accuracy. The Friedman test shows that there was a statistically significant difference between the median values of accuracy across ages at \textit{p} \textless 0.001. However, the statistical analysis indicated that the performance of the systems across ages was more consistent than their performance across questions. The results indicate that the open-ended responses from younger children were generally harder to score automatically\footnote{The full statistical analysis per age and the CD diagram per age groups are available at \url{https://github.com/venelink/mindreading-coling}}.

\begin{table}[h!]
\begin{center}
    
\begin{tabular}{|l|l|l|l|l|l|l|l|l|l|}
\hline
System              & All & 7 & 8 & 9 & 10 & 11 & 12 & 13 & 14 \\ \hline \hline
SVM (r-only)        & \textbf{.68} & .66 & .69 & .68 & .67 & .69 & .68 & .69 & .78 \\ \hline
BiLSTM (r-only)     & \textbf{.80} & .75 & .77 & .80 & .81 & .81 & .80 & .81 & .84 \\ \hline
Attention (r-only)  & \textbf{.81} & .74 & .78 & .82 & .81 & .83 & .80 & .82 & .81 \\ \hline
DistilBERT (r-only) & \textbf{.88} & .85 & .86 & .89 & .88 & .88 & .88 & .90 & .93 \\ \hline \hline
SVM (full)          & \textbf{.70} & .71 & .70 & .71 & .69 & .72 & .69 & .71 & .59 \\ \hline
BiLSTM (full)       & \textbf{.82} & .83 & .82 & .82 & .82 & .84 & .82 & .82 & .86 \\ \hline
Attention (full)    & \textbf{.83} & .83 & .80 & .83 & .83 & .84 & .82 & .83 & .91 \\ \hline
DistilBERT (full)   & \textbf{.91} & .86 & .88 & .91 & .91 & .91 & .89 & .92 & .94  \\ \hline
\end{tabular}
\caption{Comparing the System Performance by Age (Accuracy)}
\label{wmr:tab:acc-a}
\end{center}

\end{table}

\subsection{Qualitative Error Analysis}\label{wmr:error}

To understand the capabilities and limitations of the automated systems we preformed a manual analysis of a portion of the incorrect predictions for two of the systems: the BiLSTM and the DistilBERT models, trained on the ``full'' data. For each of the two models, we selected 20 incorrect predictions per question, a total of 220 incorrect predictions per model.

Our trained annotators analyzed the open-ended response for each of the selected incorrect predictions. The objective of the analysis was to determine the characteristics of the responses and to identify potential systematic sources of errors. Table \ref{wmr:tab:ql_questions} shows the characteristics used in the analysis. Each characteristic was scored in a binary ``yes''/``no'' manner to quantify the performance of the systems.

\begin{table}[h!]
\begin{center}
    
\begin{tabular}{|l|p{11.5cm}|}
\hline
Code & Characteristics \\ \hline
Acceptable & The discrepancy between human and computer rating is reasonable. (e.g.: answer could reasonably fall between 1 and 2 due to ambiguity) \\ \hline
Inappropriate & Uses mental state language (e.g. thinks, knows, pretends) that goes beyond the material or is inappropriate (e.g. in SS5 ``The burglar doesn't want to get shot''). \\ \hline
Take Voice & Writes what the character might say/think \\ \hline
Similar to (in)correct  & A correct answer which resembles an answer that is incorrect. Can also apply to an incorrect answer that resembles a correct answer.  \\ \hline
Spelling Errors & Contains spelling errors in key words that affect the overall meaning. \\ \hline
Grammar Errors & Contains grammatical errors or poor sentence construction that affect the overall meaning of the answer. \\ \hline
Long Answer & Answer is over 20 words long and there is relevant information after the 20-word cut-off point. \\ \hline
Multiple Answers & Contains at least 2 distinct answers that would be assigned different codes to one another if they appeared on their own. \\ \hline
Uncommon answer & Contains content that does not often occur in answers to this question. \\ \hline
Unclear Error & None of the other criteria can explain the error. This would be a straightforward answer for a human to code based on the current coding scheme. \\ \hline
\end{tabular}
\caption{Instructions for Qualitative Evaluation}
\label{wmr:tab:ql_questions}
\end{center}

\end{table}


Table \ref{wmr:tab:err-a} shows the result of the expert analysis of the errors. We report the percentage of incorrectly predicted responses that contain each characteristic. Since one incorrectly predicted response can have multiple characteristics, the sum of all responses exceeds 100.
We found that 14\% of the incorrect predictions made by the BiLSTM model are ``acceptable'', while for the DistilBERT model, the percentage is 23. The majority of the errors are still considered unacceptable by our experts. 

Both models have difficulty dealing with infrequent or unique responses (23\% for BiLSTM and 46\% for DistilBERT), inappropriate mentalizing (20\% for BiLSTM and 23\% for DistilBERT), and similar to (in)correct (14\% for BiLSTM and 28\% for DistilBERT). The lower performance on these types of responses indicates that the systems are potentially overfitting on the more frequent responses and are having some problems with the generalization.

Spelling mistakes (25\% for Bi-LSTM and 19\% for DistilBERT) and Grammar mistakes (9\% for BiLSTM and 11\% for DistilBERT) were also found to be a frequent source of error for both models. The high frequency of mistakes is specific to the writing of younger children. In order to improve the automatic scoring performance, we attempted to pre-process the corpus with existing spelling and grammar correcting libraries. However we did not see improvement on the task. Furthermore when manually evaluating the workings of the automatic spelling and grammar correcting tools we determined that the quality was not satisfactory in this corpus.

The length of the responses did not affect the performance of the models, even though both models used fixed length padding and truncating. However both models showed inconsistency when the participant included more than one response to the question (11\% for the BiLSTM and 17\% for DistilBERT). In this case the annotators were trained to score the ``best'' of the multiple responses.
In 12\% of the BiLSTM incorrect predictions and in 9\% of the DistilBERT ones, the experts were not able to determine any systematic source of error.

\begin{table}[h!]
\begin{center}
    
\begin{tabular}{|l|l|l|}
\hline
Characteristics & BiLSTM & DistilBERT \\ \hline
Acceptable & 14 & 23 \\ \hline
Inappropriate & 20 & 23\\  \hline
Take Voice & 0 & 4 \\ \hline
Similar to (in)correct & 14 & 28 \\ \hline
Spelling mistakes & 25 & 19 \\ \hline
Grammar mistakes & 9 & 11 \\ \hline
Long Response & 0 & 2 \\ \hline
Multiple Responses & 11 & 17 \\ \hline
Unique Response & 23 & 46 \\ \hline
Unclear Error & 12 & 9 \\ \hline

\end{tabular}
\caption{Expert Error Analysis. Frequency of Error Types (in Percentage)}
\label{wmr:tab:err-a}
\end{center}

\end{table}


\section{Discussion of the Results}\label{wmr:discussion}

In Section \ref{wmr:related} we pointed out two main limitations of the manual scoring of tests for mindreading: 1) it is labour-intensive and 2) even trained researchers may change the way they apply scoring rubrics over time. The goal of our work was to to address both of these problems.

To the best of our knowledge, this is the first work that applies NLP techniques to the automatic scoring of children's open-ended responses to experimental tests of mindreading. We demonstrated that state-of-the-art NLP systems can obtain high accuracy on the task. We also proposed a methodology for an in-depth evaluation and error analysis. We showed that while the overall accuracy and macro-F1 are very high, the performance varies significantly per question. The automated systems also faced problems when the answers were more creative and unique. The performance of the systems also dropped due to spelling and grammar errors. The in-depth evaluation and error analysis gives clear directions for improvement before an automated solution can be implemented in practice.

This work, and in particular the expert error analysis also helped to re-evaluate and improve the original tasks. During the qualitative analysis trained test raters developed suggestions on how to improve the existing guidelines for human annotators. This will improve the consistency of the human annotation in further studies\footnote{After modifying the annotation guidelines, we re-annotated the whole corpus. Both the original and the re-annotated version of the corpus are available. The re-annotated corpus also contains responses from 77 additional participants, that were processed after the submission deadline. The results reported in the paper are based on the original corpus.}.

This paper emphasizes the advantages of inter-disciplinary work and shows the importance of incorporating in-domain experts (in our case developmental psychologists) when designing and implementing a novel NLP solution. The collaboration between researchers from different fields provides different perspectives on the problem. It benefits the formalization of the task and the evaluation of the systems.

\section{Conclusions and Future Work}\label{wmr:conclusions}

In this paper we presented the first NLP approach for the automated scoring of mindreading in middle childhood and early adolescence. An inter-disciplinary team of psychologists, computer scientists, and linguists created MIND-CA - a corpus of 11,311 responses from 1,066 children aged from 7 to 14. We trained and evaluated four different Machine Learning systems on the data and obtained promising results. We presented a novel methodology for evaluation and error analysis inspired by the inter-disciplinary collaboration between psychologists and NLP researchers. 

In future work we want to move towards designing a web application that can be used to gather the responses from the children and score them automatically in an end-to-end manner.
To ensure the quality of an end-to-end solution and the reliability of the automated scoring, we will further test the capabilities of the automated scoring systems and ensure that they can generalize well to novel data. We are in the process of digitalizing and scoring the responses of  1,000 more 8- to 13-year-old children, gathered in a different time span and from different schools than MIND-CA. This way we will double the existing corpus size and will also be able to test the performance of the systems on further, unseen data.

We are also exploring the possibilities to use data augmentation on the current corpus. Data augmentation can be used to address the problem of imbalanced distribution of labels for some of the questions. In the experimental setup we plan to incorporate more extra-linguistic data into the evaluation, for example the gender of the participants and their responses to standardized language ability tests and psychopathology.

Finally, the analysis of the performance of the systems showed that spelling and grammar mistakes were a systematic source of error and the out-of-the-box correction systems were not well adapted to deal with children's open-ended text responses. We will examine the possibility of developing a spelling correction system designed to work with participants in middle childhood and early adolescence. 





\section*{Acknowledgments}

This project was funded by a grant from Wellcome to R. T. Devine. We also want to thank the anonymous reviewers for their feedback and suggestions.



\bibliographystyle{coling}
\bibliography{coling2020}

\end{document}